\documentclass{article}
\usepackage{ijcai18}

\usepackage{times}
\usepackage{xcolor}
\usepackage{soul}
\usepackage[utf8]{inputenc}
\usepackage[small]{caption}

\usepackage{graphicx}
\usepackage{arydshln}
\usepackage{multirow} 
\usepackage{booktabs}
\usepackage{bm}
\usepackage{epsfig}
\usepackage{amsmath}
\usepackage{amssymb}
\usepackage{hyperref}

\title{Mixed Link Networks}
\author{Anonymous}

\author{
	Wenhai Wang$^{*1}$, 
	Xiang Li\thanks{Authors contributed equally}$^{2}$, 
	Jian Yang$^2$,
	Tong Lu$^{1}$
	\\ 
	$^1$ National Key Lab for Novel Software Technology, Nanjing University \\
	$^2$ DeepInsight@PCALab, Nanjing University of Science and Technology \\
	wangwenhai362@163.com,
	xiang.li.implus@njust.edu.cn,
	csjyang@njust.edu.cn,
	lutong@nju.edu.cn,
}

\begin{document}
	\maketitle	
	\begin{abstract}
		Basing on the analysis by revealing the equivalence of modern networks, we find that both ResNet and DenseNet are essentially derived from the same ``dense topology'', yet they only differ in the form of connection -- addition (dubbed ``inner link'') \emph{vs.} concatenation (dubbed ``outer link''). However, both two forms of connections have the superiority and insufficiency. To combine their advantages and avoid certain limitations on representation learning, we present a highly efficient and modularized Mixed Link Network (MixNet) which is equipped with flexible inner link and outer link modules. Consequently, ResNet, DenseNet and Dual Path Network (DPN) can be regarded as a special case of MixNet, respectively. Furthermore, we demonstrate that MixNets can achieve superior efficiency in parameter over the state-of-the-art architectures on many competitive datasets like CIFAR-10/100, SVHN and ImageNet.
	\end{abstract}
	
	\section{Introduction}
	\label{sec:intro}
	The exploration of connectivity patterns of deep neural networks has attracted extensive attention in the literature of Convolutional Neural Networks (CNNs). LeNet \cite{lecun1998gradient} originally demonstrated its \emph{layer-wise feed-forward} pipeline, and later GoogLeNet \cite{szegedy2015going} introduced more effective \emph{multi-path} topology. Recently, ResNet \cite{he2016deep,he2016identity} successfully adopted \emph{skip connection} which transferred early information through identity mapping by element-wisely adding input features to its block outputs. DenseNet \cite{huang2016densely} further proposed a seemingly ``different'' topology by using \emph{densely connected path} to concatenate all the previous raw input features with the output ones.
	
	\begin{figure}
		\vspace{-10pt}
		\centering
		\setlength{\fboxrule}{0pt}
		\fbox{\includegraphics[width=0.4\textwidth]{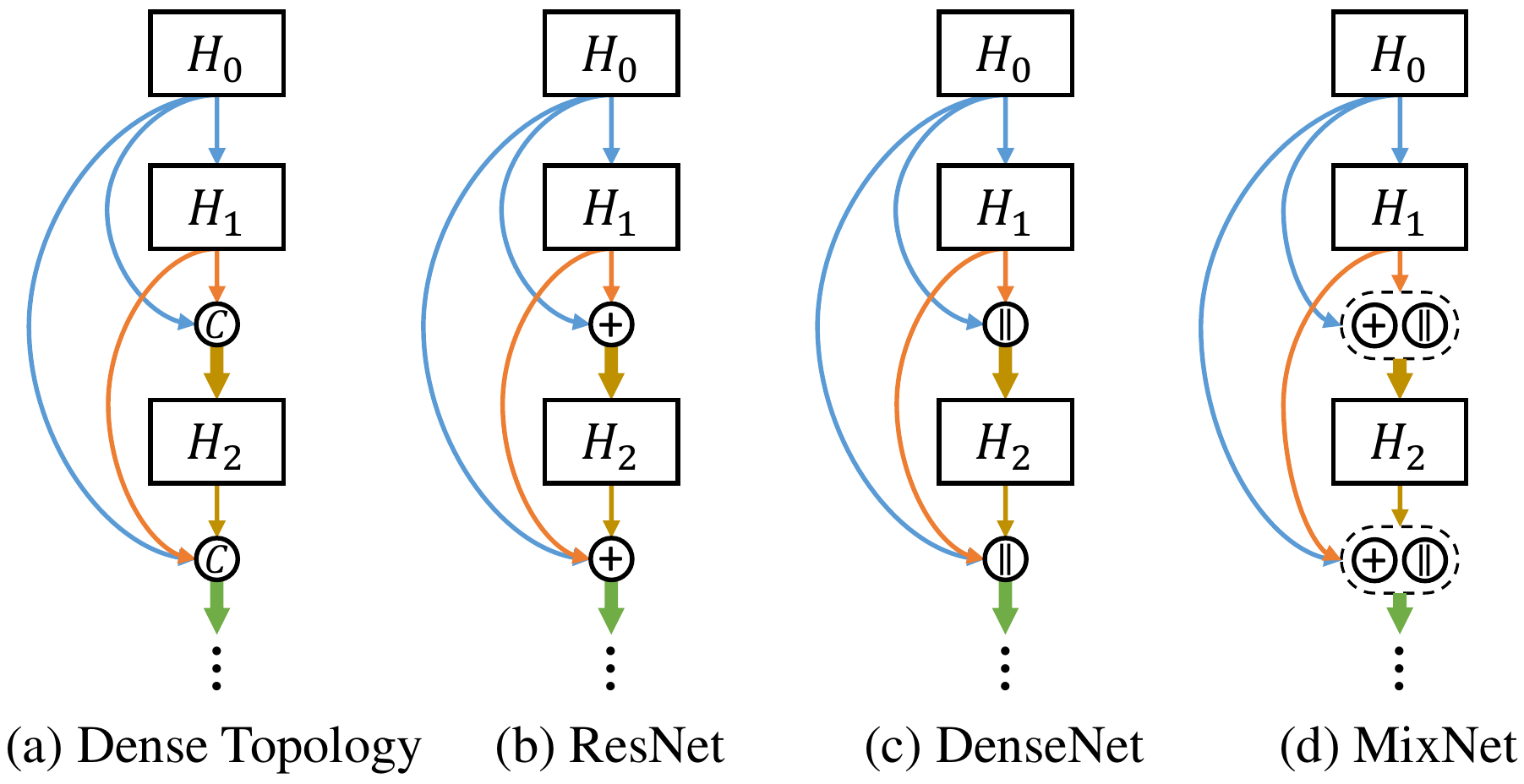}}
		\vspace{-8pt}
		\caption{{The topological relations of different types of neural networks. The symbols ``$+$'' and ``$\parallel$'' denote element-wise addition and concatenation, respectively. (a) shows the general form of ``dense topology''. $C(\cdot)$ refers to the connection function. (b) shows ResNet in the perspective of ``dense topology''. (c) shows the path topology of DenseNet. (d) shows the path topology of MixNet.}}
		\label{fig:diff-nn}
		\vspace{-10pt}
	\end{figure}
	For the two recent ResNet and DenseNet, despite their \emph{externally large} difference in path topology (\emph{skip connection vs. densely connected path}), \textbf{we discover and prove that both of them are essentially derived from the \emph{same} ``dense topology''} (Fig. \ref{fig:diff-nn} (a)), where their only difference lies in the form of connection (``$+$'' in Fig. \ref{fig:diff-nn} (b) \emph{vs.} ``$\parallel$'' in Fig. \ref{fig:diff-nn} (c)). Here, ``dense topology'' is defined as a path topology in which each layer $H_\ell$ is connected with all the previous layers $H_0, H_1, ..., H_{\ell - 1}$ using the connection function $C(\cdot)$ . The great effectiveness of ``dense topology'' has been proved via the significant success of both ResNet and DenseNet, yet the form of connection in ResNet and DenseNet still has room for improvement. For example, too many additions on the same feature space may impede the information flow in ResNet \cite{huang2016densely}, and there may be the same type of raw features from different layers, which leads to a certain redundancy in DenseNet \cite{chen2017dual}.  Therefore, the question ``does there exist a more efficient form of connection in the dense topology'' still remains to be further explored. 
	
	
	To address the problem, in this paper, we propose a novel Mixed Link Network (MixNet) with an efficient form of connection (Fig. \ref{fig:diff-nn} (d)) in the ``dense topology''. That is, we mix the connections in ResNet and DenseNet, in order to combine both the advantages of them and avoid their possible limitations. In particular, the proposed MixNets are equipped with both inner link modules and outer link modules, where an inner link module refers to \emph{additive} feature vectors (similar connection in ResNet), while an outer link module stands for \emph{concatenated} ones (similar connection in DenseNet). More importantly, in the architectures of MixNets, these two types of link modules are flexible with their positions and sizes. As a result, ResNet, DenseNet and the recently proposed Dual Path Network (DPN) \cite{chen2017dual} can be regarded as a special case of MixNet, respectively (see the details in Fig.~\ref{fig:archs} and Table \ref{tab: archs}). 
	
	To show the efficiency and effectiveness of the proposed MixNets, we conduct extensive experiments on four competitive benchmark datasets, namely, CIFAR-10, CIFAR-100, SVHN and ImageNet. The proposed MixNets require fewer parameters than the existing state-of-the-art architectures whilst achieving better or at least comparable results. Notably, on CIFAR-10 and CIFAR-100 datasets, MixNet-250 surpasses ResNeXt-29 (16$\times$64d) with 57\% less parameters. On ImagNet dataset, the results of MixNet-141 are comparable to the ones of DPN-98 with 50\% fewer parameters.
	
	{The main contributions of this paper are as follows:}
	\begin{itemize}
		\setlength{\itemsep}{1pt}
		\setlength{\parsep}{1pt}
		\setlength{\parskip}{1pt}
		\item ResNet and DenseNet are proved to have  the \emph{same} path topology -- ``dense topology'' essentially, whilst their only difference lies in the form of connections.
		\item A highly modularized Mixed Link Network (MixNet) is proposed, which has a more efficient connection -- the blending of flexible inner link modules and outer link modules.
		\item The relation between MixNet and modern networks (ResNet, DenseNet and DPN) is discussed, and these modern networks are shown to be specific instances of MixNets.
		\item MixNet demonstrates its superior efficiency in parameter over the state-of-the-art architectures on many competitive benchmarks. 
	\end{itemize}
	
	

	\section{Related Work}
	\label{sec:rel-work}
	Designing effective path topologies always pushes the frontier of the advanced neural network architecture. Following the initial layer-wise feed-forward pipeline \cite{lecun1998gradient}, AlexNet \cite{lecun1998gradient} and VGG \cite{simonyan2014very} showed that building deeper networks with tiny convolutional kernels is a promising way to increase the learning capacity of neural network. GoogLeNet \cite{szegedy2015going} demonstrated that a multi-path topology (codenamed Inception) could easily outperform previous feed-forward baselines by blending various information flows. The effectiveness of multi-path topology was further validated in FractalNet \cite{larsson2016fractalnet}, Highway Networks \cite{srivastava2015training}, DFN \cite{wang2016deeply}, DFN-MR \cite{zhao2016connection}, and IGC \cite{zhang2017interleaved}. A recurrent connection topology \cite{liang2015recurrent} was proposed to integrate the context information. Perhaps the most revolutionary topology -- skip connection was successfully adopted by ResNet \cite{he2016deep,he2016identity}, where micro-blocks were built sequentially and the skip connection bridged the micro-block's input features with its output ones via identity mappings. Since then, different works based on ResNet have arisen, aiming to find a more efficient transformer of that micro-block, such as WRN \cite{zagoruyko2016wide}, Multi-ResNet \cite{abdi2016multi} and ResNeXt \cite{xie2017aggregated}. Furthermore, DenseNet \cite{huang2016densely} achieved comparable accuracy with deep ResNet by proposing the densely connected topology, which connects each layer to its previous layers by concatenation. Recently, DPN \cite{chen2017dual} directly combines the two paths -- ResNet path and DenseNet path together by a shared feature embedding in order to enjoy a mutual improvement.    
	
	
	\section{Dense topology}
	
	\label{sec:res-dense-dsn} 
	In this section, we first introduce and formulate the ``dense topology''.
	We then prove that both ResNet and DenseNet are intrinsically derived from the same ``dense topology'', and they only differ in the specific form of connection (addition \emph{vs.} concatenation).
	Furthermore, we present analysis on strengths and weaknesses of these two network architectures, which motivates us to develop Mixed Link Networks.
	
\begin{figure}
		\vspace{-10pt}
		\centering
		\setlength{\fboxrule}{0pt}
		\fbox{\includegraphics[width=0.42\textwidth]{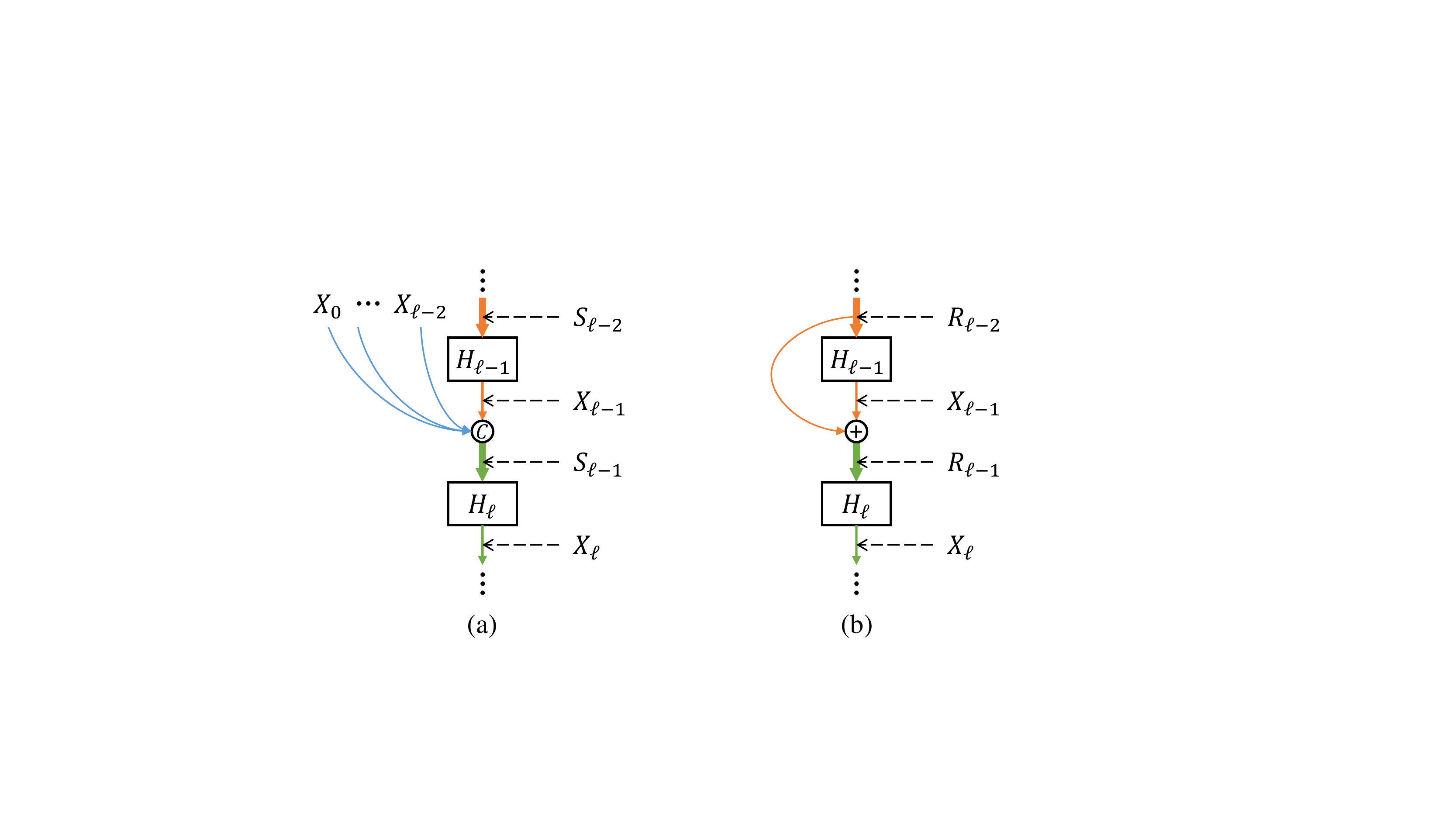}}
		\vspace{-8pt}
		\caption{The key annotations for $H(\cdot)$, $X$, $S$ and $R$.}
		\label{fig:connection}
		\vspace{-10pt}
	\end{figure}

	\textbf{Definitions of ``dense topology''.} Let us consider 
	a network that comprises $L$ layers, each of which implements a non-linear transformation $H_\ell(\cdot)$, where $\ell$ indexes the layer. $H_\ell(\cdot)$ could be a composite function of several operations such as linear transformation, convolution, activation function, pooling \cite{lecun1998gradient}, batch normalization \cite{ioffe2015batch}. As illustrated in Fig.~\ref{fig:connection} (a), $X_\ell$ refers to the immediate output of the transformation $H_\ell(\cdot)$ and $S_\ell$ is the result of the connection function $C(\cdot)$ whose inputs come from all the previous feature-maps $X$ (i.e., $X_0, X_1, ..., X_\ell$). Initially, $S_0$ equals $X_0$. As mentioned in Section \ref{sec:intro}, ``dense topology'' is defined as a path topology where each layer is connected with all the previous layers. Therefore, we can formulate the general form of ``dense topology'' simply as:
	\begin{equation}
	X_\ell = H_\ell(C(X_0, X_1, \cdots, X_{\ell - 1})).
	\label{eqn:dense-topo}
	\end{equation}
	
	\textbf{DenseNet is derived from ``dense topology'' obviously.} For DenseNet \cite{huang2016densely}, the input of $\ell^{th}$ layer is the concatenation of the outputs $X_0, X_1, ..., X_{\ell - 1}$ from all the preceding layers. Therefore, we can write DenseNet as: 
	\begin{equation}
	X_\ell = H_\ell(X_0 \parallel X_1 \parallel \cdots \parallel X_{\ell - 1}),
	\label{eqn:densenet}
	\end{equation}
	where ``$\parallel$'' refers to the concatenation. As shown in Eqn. \eqref{eqn:dense-topo} and Eqn. \eqref{eqn:densenet}, DenseNet directly follows the formulation of ``dense topology'', whose connection function is the pure concatenation (Fig. \ref{fig:diff-nn} (c)).
	
	\textbf{ResNet is also derived from ``dense topology''.} We then explain that ResNet also follows the ``dense topology'' whose connection is accomplished by addition. Given the standard definition from \cite{he2016identity}, ResNet poses a skip connection that bypasses the non-linear transformations $H_\ell(\cdot)$ with an identity mapping as: 
	\begin{equation}
	R_\ell = H_\ell(R_{\ell - 1}) + R_{\ell - 1},
	\label{eqn:resnet0}
	\end{equation}
	where $R$ refers to the feature-maps directly after the skip connection (Fig. \ref{fig:connection} (b)). Initially, $R_0$ equals $X_0$. Now we concentrate on $X_\ell$ which is the output of $H_\ell(\cdot)$ as well:
	\begin{equation}
	X_\ell = H_\ell(R_{\ell - 1}).
	\label{eqn:resnet1}
	\end{equation}
	
	By substituting Eqn. \eqref{eqn:resnet0} into Eqn. \eqref{eqn:resnet1} recursively, we can rewrite Eqn. \eqref{eqn:resnet1} as:
	\begin{eqnarray}
	X_\ell & = & H_\ell(R_{\ell - 1}) \nonumber = H_\ell(H_{\ell - 1}(R_{\ell - 2}) + R_{\ell - 2}) \nonumber \\
	& = & H_\ell(H_{\ell - 1}(R_{\ell - 2}) + H_{\ell - 2}(R_{\ell - 3}) + R_{\ell - 3}) \nonumber \\
	& = & \cdots \nonumber \\
	& = & H_\ell(\sum\nolimits_{i = 1}^{\ell - 1}{H_{i}(R_{i - 1})} + R_0) \nonumber \\
	& = & H_\ell(\sum\nolimits_{i = 1}^{\ell - 1}{X_i}+ X_0) \nonumber \\
	& = & H_\ell(X_0 + X_1 + \cdots + X_{\ell - 1}).
	\label{eqn:resnet2}
	\end{eqnarray}
	
	As shown in Eqn. \eqref{eqn:resnet2} clearly, $R_{\ell - 1}$ in ResNet is deduced to be the element-wise addition result of all the previous layers -- $X_0, X_1, ..., X_{\ell - 1}$. It proves that ResNet is actually identical to a form of ``dense topology'', where the connection function $C(\cdot)$ is specified to the addition (Fig. \ref{fig:diff-nn} (b)). 
	
	The above analyses reveal that ResNet and DenseNet share the same ``dense topology'' in essence. Therefore, the ``dense topology'' is confirmed to be a \emph{fundamental} and \emph{significant} path topology that works practically, due to the extraordinary effectiveness of both ResNet and DenseNet in the recent progress. Meanwhile, from Eqn. \eqref{eqn:densenet} and Eqn. \eqref{eqn:resnet2}, the only difference between ResNet and DenseNet is the connection function $C(\cdot)$ (``$+$'' \emph{vs.} ``$\parallel$'') obviously. 
	
	\textbf{Analysis of ResNet.} The connection in ResNet is \emph{only} the additive form (``$+$'') that operates on the entire feature map. It combines the features from previous layers by element-wise addition, which makes the features more expressive and eases the gradient flow for optimization simultaneously. However, too many additions on the same feature space may impede the information flow in the network \cite{huang2016densely}, which motivates us to develop a ``shifted additions'', by dislocating/shifting the additive positions in subsequent feature spaces along multiple layers (e.g., the black arrow in Fig. \ref{fig:archs} (e)), to alleviate this problem.
	
	\textbf{Analysis of DenseNet.} The connection in DenseNet is \emph{only} the concatenative connection (``$\parallel$'') which increases the feature dimension gradually along the depths. It concatenates the raw features from previous layers to form the input of the new layer. Concatenation allows the new layer to receive the raw features directly from previous layers and it also improves the flow of information between layers. However, there may be the same type of features from different layers, which leads to a certain redundancy \cite{chen2017dual}. This limitation also inspires us to introduce the ``shifted additions'' (e.g., the black arrow in Fig. \ref{fig:archs} (e)) on these raw features in purpose of a modification to avoid that redundancy to some extent. 
	
	
	\begin{figure}
		\centering
		\setlength{\fboxrule}{0pt}
		\fbox{\includegraphics[width=0.46\textwidth]{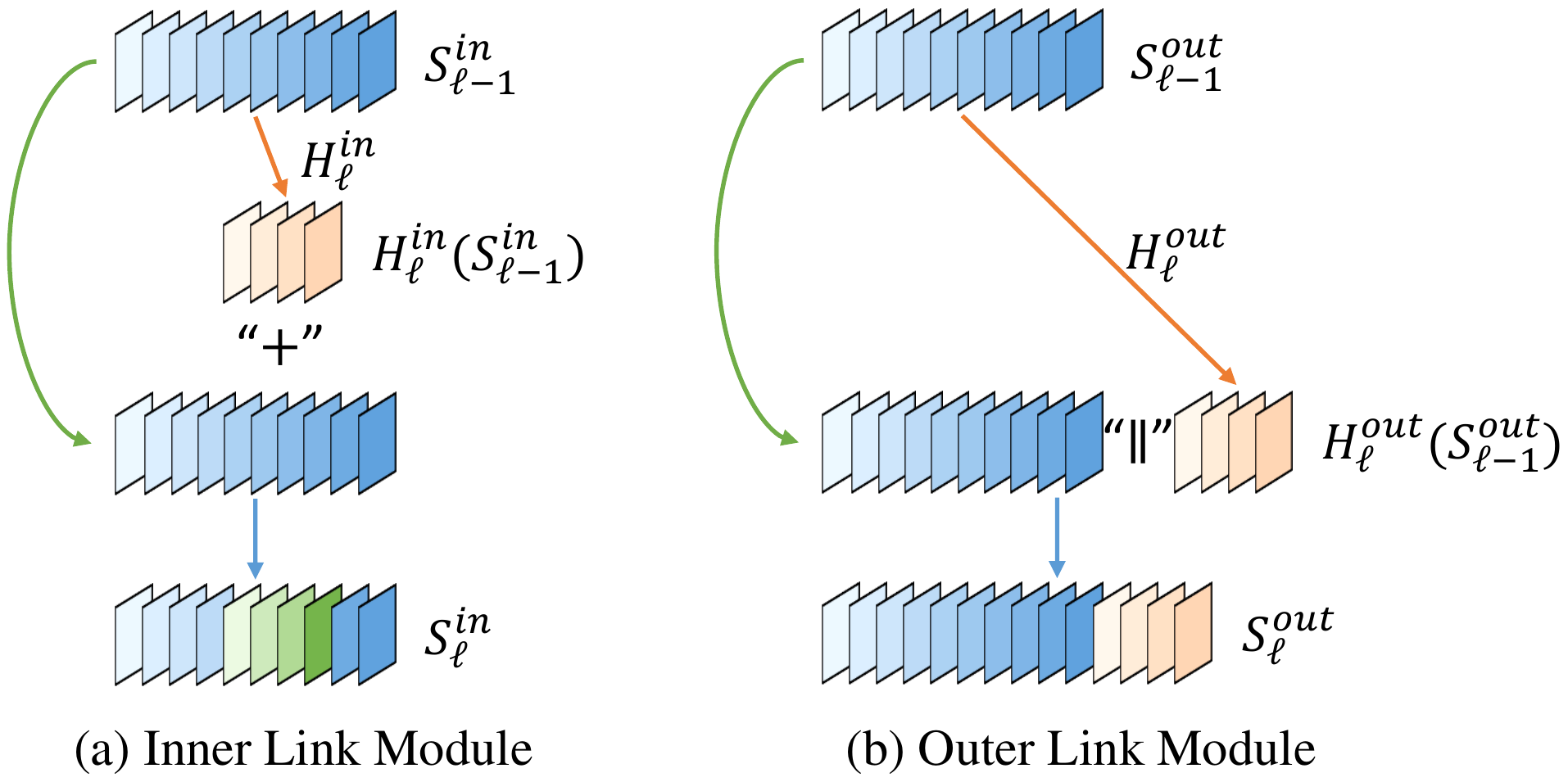}}
		\vspace{-8pt}
		\caption{The examples of inner/outer link module. The symbol ``$+$'' and ``$\parallel$'' denote addition and concatenation, respectively. The green arrows refer to duplication operation. (a) and (b) show the examples of inner link module and outer link module, respectively.}
		\label{fig:stack-in-stack-out}
		\vspace{-10pt}
	\end{figure}
	
	\section{Mixed Link Networks}
	\label{sec:DSCN}
	In this section, we first introduce and formulate the inner/outer link modules. Next, we present the generalized mixed link architecture with flexible inner/outer link modules and propose Mixed Link Network (MixNet), a representative form of the generalized mixed link architecture. At last, we describe the implementation details of MixNets.
	
	\subsection{Inner/Outer Link Module}
	The inner link modules are based on the additive connections. Following the above preliminaries, we denote the output $S^{in}_\ell$ which contains the inner link part as\footnote{Please note that we omit the possible positional parameters to align/place the inner link parts $H_\ell^{in}(S^{in}_{\ell - 1})$ for simplicity, and it will be discussed in the following subsection (Fig.~\ref{fig:archs} and Table \ref{tab: archs}).}: 
	\begin{eqnarray}
	S^{in}_\ell & = & \sum\nolimits_{i=0}^{\ell} X_i =  S^{in}_{\ell - 1} + X_\ell \nonumber \\
	& = & S^{in}_{\ell - 1} + H_\ell^{in}(S^{in}_{\ell - 1}),
	\label{eqn:stack-in}
	\end{eqnarray} 
	where $H^{in}_\ell(\cdot)$ refers to the function of producing feature-maps for inner linking -- element-wisely adding new features $H_\ell^{in}(S^{in}_{\ell - 1})$ inside the original ones $S^{in}_{\ell - 1}$ (Fig. \ref{fig:stack-in-stack-out} (a)). 
	
	The outer link modules are based on the concatenated connection. Similarly, we have $S^{out}_\ell$ as: 
	\begin{eqnarray}
	S^{out}_\ell & = & X_0 \parallel X_1 \parallel \cdots \parallel X_\ell = S^{out}_{\ell - 1} \parallel X_\ell \nonumber \\
	& = & S^{out}_{\ell - 1} \parallel H_\ell^{out}(S^{out}_{\ell - 1}),
	\label{eqn:stack-out}
	\end{eqnarray}
	where $H^{out}_\ell(\cdot)$ refers to the function of producing feature-maps for outer linking -- appending new features $H_\ell^{out}(S^{out}_{\ell - 1})$ outside the original ones $S^{out}_{\ell - 1}$ (Fig. \ref{fig:stack-in-stack-out} (b)).
	
	\begin{figure}
		\centering
		\setlength{\fboxrule}{0pt}
		\fbox{\includegraphics[width=0.252\textwidth]{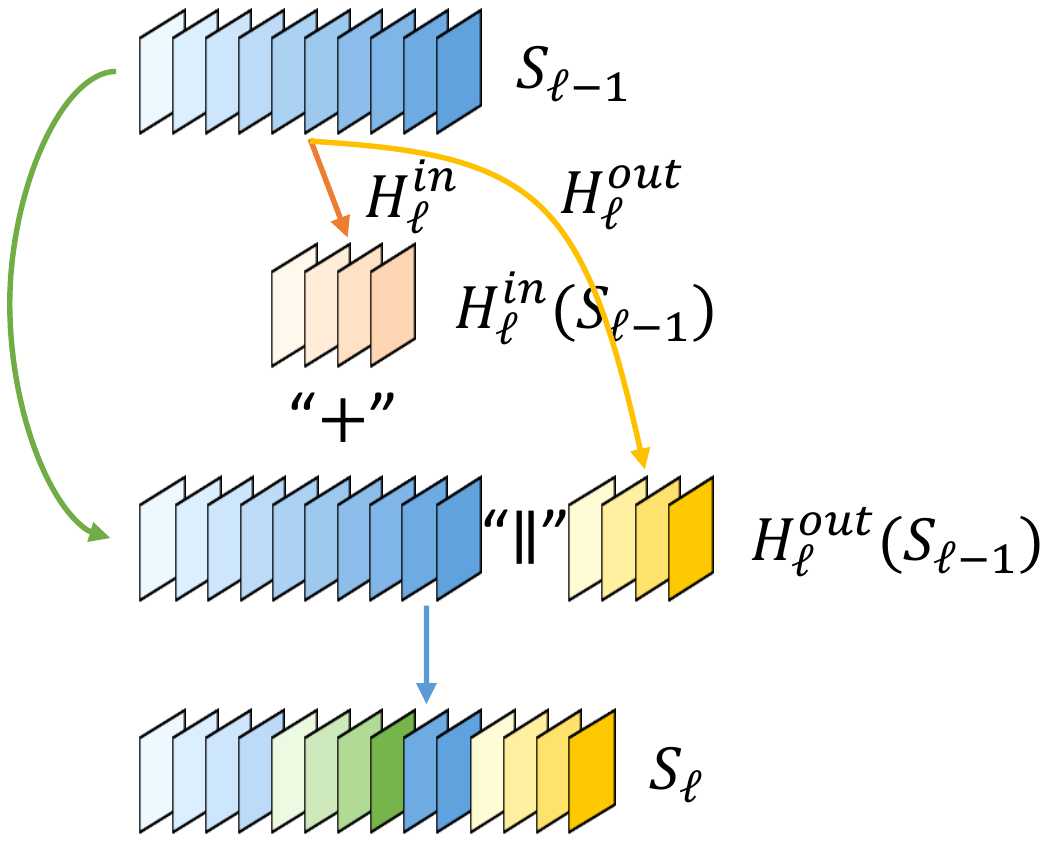}}
		\vspace{-8pt}
		\caption{The example of mixed link architecture. The symbol ``$+$'' and ``$\parallel$'' represent addition and concatenation, respectively. The green arrows denote duplication operation.}
		\label{fig:stack-in-out}
		\vspace{-10pt}
	\end{figure}
	
	\subsection{Mixed Link Architecture}
	Basing on the analyses in Section \ref{sec:res-dense-dsn}, we introduce the mixed link architecture which embraces both inner link modules and outer link modules (Fig. \ref{fig:stack-in-out}). The mixed link architecture can be formulated as Eqn. \eqref{eqn:stack-in-out}, a flexible combination of Eqn. \eqref{eqn:stack-in} and Eqn. \eqref{eqn:stack-out}, to get a blending feature output $S_\ell$:
	\begin{equation}
	S_\ell = (S_{\ell - 1} + H^{in}_\ell(S_{\ell - 1})) \parallel H^{out}_\ell(S_{\ell - 1}).
	\label{eqn:stack-in-out}
	\end{equation}
	
\begin{figure}
		\centering
		\setlength{\fboxrule}{0pt}
		\fbox{\includegraphics[width=0.46\textwidth]{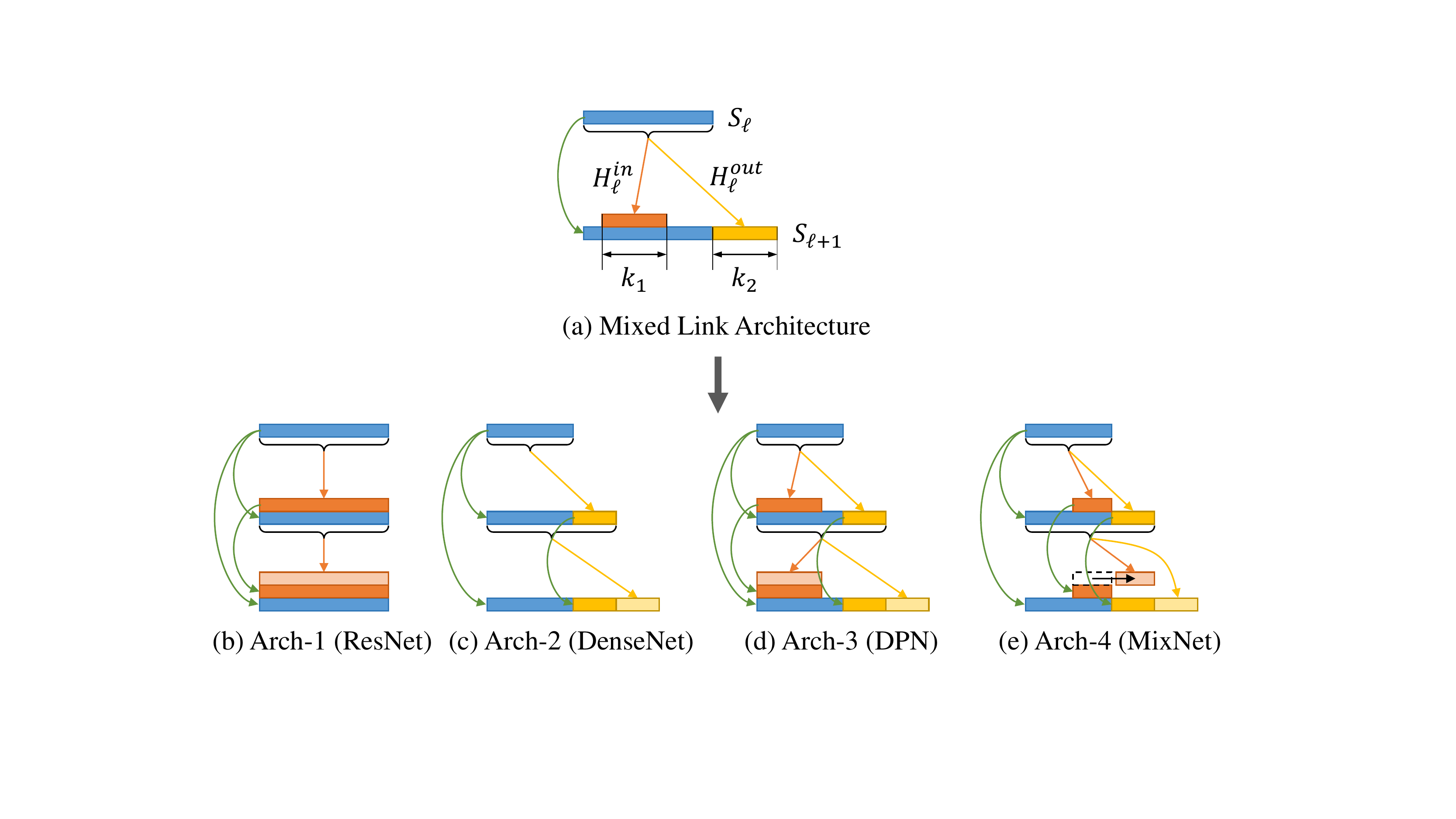}}
		\vspace{-8pt}
		\caption{Four architectures derived from mixed link architecture. The view is placed on the channels of one location of feature-maps in convolutional neural networks. The orange arrows denote the function $H^{in}_\ell(\cdot)$ for the inner link module. The yellow arrows denote the function $H^{out}_\ell(\cdot)$ for the outer link module. The green arrows refer to duplication operation. The vertically aligned features are merged by element-wise addition, and the horizontally aligned features are merged by concatenation. (a) shows the generalized mixed link architecture. (b), (c), (d) and (e) are the four derivative architectures with various representative settings of (a).}
		\label{fig:archs}
		\vspace{-10pt}
	\end{figure}

	\textbf{Definitions of parameters ($k_1$, $k_2$, fixed/unfixed) for mixed link architecture.} Here we denote the channel number of feature-maps produced by $H^{in}_\ell(\cdot)$ and $H^{out}_\ell(\cdot)$ as $k_1$ and $k_2$, respectively. That is, $k_1$ is the inner link size for inner link modules, and $k_2$ controls the outer link size for outer link modules. As for the positional control for inner link modules, we simplify it into two choices -- ``fixed'' or ``unfixed''. The ``fixed'' term is easy to understand -- all the features are merged together by addition over the same fixed space, as in ResNet. Here is the explanation for ``unfixed'': there are extremely exponential combinations to pose the inner link modules' positions along multiple layers, and learning the variable position is currently unavailable since their arrangement is not derivable directly. Therefore, we make a compromise and choose one simple series of the unfixed-position version -- the ``shifted addition'' (Fig. \ref{fig:archs} (e)) as mentioned in our motivations in Section \ref{sec:res-dense-dsn}. Specifically, the position of inner link part exactly aligns with the growing boundary of entire feature embedding (see the black arrow in Arch-4) when the outer link parts increase the overall feature dimension. We denote this Arch-4 (Fig. \ref{fig:archs} (e)) to be our proposed model exactly -- Mixed Link Network (MixNet). In summary, we have defined the above two simple options for controlling the positions of inner link modules as {-- ``fixed'' and ``unfixed'' }.
	
	\textbf{Modern networks are special cases of MixNets.} It can be seen from Fig. \ref{fig:archs} that the mixed link architecture (Fig. \ref{fig:archs} (a)) with different parametric configurations can reach four representative architectures (Fig. \ref{fig:archs} (b)(c)(d)(e)). The configurations of these corresponding architectures are listed in Table \ref{tab: archs}. We show that MixNet is a more generalized form than other exsiting modern networks, under the perspective of mixed link architecture. Therefore, ResNet, DenseNet and DPN can be treated as a specific instance of MixNets, respectively. 
	
	\begin{table}
		\scriptsize
		\centering
		\renewcommand\arraystretch{1.4}
		\newcommand{\tabincell}[2]{\begin{tabular}{@{}#1@{}}#2\end{tabular}}
		\caption{The configurations of the four representative architectures.}
		\vspace{-8pt}
		\begin{tabular}{l|c|c}
			\hline
			Architecture & \tabincell{c}{Inner Link  Module Setting} & \tabincell{c}{Outer Link Module Setting} \\
			\hline
			Arch-1 \tiny{(ResNet)} & $k_1 > 0$, fixed & $k_2 = 0$ \\
			\hline
			Arch-2 \tiny{(DenseNet)} & $k_1 = 0$ & $k_2 > 0$ \\
			\hline
			Arch-3 \tiny{(DPN)} & $k_1 > 0$, fixed & $k_2 > 0$ \\
			\hline
			Arch-4 \tiny{(MixNet)}& $k_1 > 0$, unfixed & $k_2 > 0$ \\
			\hline
		\end{tabular}
		\label{tab: archs}
		\vspace{-10pt}
	\end{table}
	
	\subsection{Implementation Details of MixNets}
	\label{sec:net-config}
	\begin{table*}
		\scriptsize
		\centering
		\renewcommand\arraystretch{1.2}
		\newcommand{\tabincell}[2]{\begin{tabular}{@{}#1@{}}#2\end{tabular}}
		\caption{MixNet architectures for ImageNet. $k_1$ and $k_2$ denote the parameters for inner and outer link modules, respectively.}
		\vspace{-10pt}
		\scalebox{1}{
			\begin{tabular}{c|c|c|c|c}
				\hline
				Layers & \tabincell{c}{Output \\ Size} & \tabincell{c}{MixNet-105 \\ ($k_1 = 32, k_2 = 32$)} & \tabincell{c}{MixNet-121 \\ ($k_1 = 40, k_2 = 40$)} & \tabincell{c}{MixNet-141 \\ ($k_1 = 48, k_2 = 48$)} \\
				\hline
				Convolution & $112 \times 112$ &
				\multicolumn{3}{c}{$7 \times 7$ conv, stride 2} \\
				\hline
				Pooling & $56 \times 56$ &
				\multicolumn{3}{c}{$3 \times 3$ max pool, stride 2} \\
				\hline
				\tabincell{c}{Mixed Link \\ Block (1)} & $56 \times 56$ &
				$\begin{bmatrix}
				\begin{bmatrix}
				\begin{array}{l}
				1 \times 1\mbox{, conv} \\
				3 \times 3\mbox{, conv} \\
				\end{array}
				\end{bmatrix} \times 2
				\end{bmatrix} \times 6$
				&
				$\begin{bmatrix}
				\begin{bmatrix}
				\begin{array}{l}
				1 \times 1\mbox{, conv} \\
				3 \times 3\mbox{, conv} \\
				\end{array}
				\end{bmatrix} \times 2
				\end{bmatrix} \times 6$
				&
				$\begin{bmatrix}
				\begin{bmatrix}
				\begin{array}{l}
				1 \times 1\mbox{, conv} \\
				3 \times 3\mbox{, conv} \\
				\end{array}
				\end{bmatrix} \times 2
				\end{bmatrix} \times 6$ \\
				\hline
				Convolution & $56 \times 56$ &
				\multicolumn{3}{c}{$1 \times 1$ conv} \\
				\hline
				Pooling & $28 \times 28$ &
				\multicolumn{3}{c}{$2 \times 2$ average pool, stride 2} \\
				\hline
				\tabincell{c}{Mixed Link \\ Block (2)} & $28 \times 28$ &
				$\begin{bmatrix}
				\begin{bmatrix}
				\begin{array}{l}
				1 \times 1\mbox{, conv} \\
				3 \times 3\mbox{, conv} \\
				\end{array}
				\end{bmatrix} \times 2
				\end{bmatrix} \times 12$
				&
				$\begin{bmatrix}
				\begin{bmatrix}
				\begin{array}{l}
				1 \times 1\mbox{, conv} \\
				3 \times 3\mbox{, conv} \\
				\end{array}
				\end{bmatrix} \times 2
				\end{bmatrix} \times 12$
				&
				$\begin{bmatrix}
				\begin{bmatrix}
				\begin{array}{l}
				1 \times 1\mbox{, conv} \\
				3 \times 3\mbox{, conv} \\
				\end{array}
				\end{bmatrix} \times 2
				\end{bmatrix} \times 12$ \\
				\hline
				Convolution & $56 \times 56$ &
				\multicolumn{3}{c}{$1 \times 1$ conv} \\
				\hline
				Pooling & $28 \times 28$ &
				\multicolumn{3}{c}{$2 \times 2$ average pool, stride 2} \\
				\hline
				\tabincell{c}{Mixed Link \\ Block (3)} & $14 \times 14$ &
				$\begin{bmatrix}
				\begin{bmatrix}
				\begin{array}{l}
				1 \times 1\mbox{, conv} \\
				3 \times 3\mbox{, conv} \\
				\end{array}
				\end{bmatrix} \times 2
				\end{bmatrix} \times 20$
				&
				$\begin{bmatrix}
				\begin{bmatrix}
				\begin{array}{l}
				1 \times 1\mbox{, conv} \\
				3 \times 3\mbox{, conv} \\
				\end{array}
				\end{bmatrix} \times 2
				\end{bmatrix} \times 24$
				&
				$\begin{bmatrix}
				\begin{bmatrix}
				\begin{array}{l}
				1 \times 1\mbox{, conv} \\
				3 \times 3\mbox{, conv} \\
				\end{array}
				\end{bmatrix} \times 2
				\end{bmatrix} \times 30$ \\
				\hline
				Convolution & $56 \times 56$ &
				\multicolumn{3}{c}{$1 \times 1$ conv} \\
				\hline
				Pooling & $28 \times 28$ &
				\multicolumn{3}{c}{$2 \times 2$ average pool, stride 2} \\
				\hline
				\tabincell{c}{Mixed Link \\ Block (4)} & $7 \times 7$ &
				$\begin{bmatrix}
				\begin{bmatrix}
				\begin{array}{l}
				1 \times 1\mbox{, conv} \\
				3 \times 3\mbox{, conv} \\
				\end{array}
				\end{bmatrix} \times 2
				\end{bmatrix} \times 12$
				&
				$\begin{bmatrix}
				\begin{bmatrix}
				\begin{array}{l}
				1 \times 1\mbox{, conv} \\
				3 \times 3\mbox{, conv} \\
				\end{array}
				\end{bmatrix} \times 2
				\end{bmatrix} \times 16$
				&
				$\begin{bmatrix}
				\begin{bmatrix}
				\begin{array}{l}
				1 \times 1\mbox{, conv} \\
				3 \times 3\mbox{, conv} \\
				\end{array}
				\end{bmatrix} \times 2
				\end{bmatrix} \times 20$ \\
				\hline
				\multirow{2}{*}{\tabincell{c}{Classification \\ Layer}} & $1 \times 1$ & 
				\multicolumn{3}{c} {$7 \times 7$ global average pool} \\
				\cline{2-5}
				& 1000 &
				\multicolumn{3}{c}{1000D fully-connected, softmax} \\
				\hline
		\end{tabular}}
		\label{tab: nn-arch}
		\vspace{-10pt}
	\end{table*}
	
	The proposed network consists of multiple mixed link blocks. Each mixed link block has several layers, whose structure follows Arch-4 (Fig. \ref{fig:archs} (e)). Motivated from the common practices \cite{szegedy2016rethinking,he2016deep}, we introduce bottleneck layers as unitary elements in MixNets. That is, we implement both $H^{in}_\ell(\cdot)$ and $H^{out}_\ell(\cdot)$ with such a bottleneck layer -- BN-ReLU-Conv(1, 1)-BN-ReLU-Conv(3, 3). Here BN, ReLU, and Conv refer to batch normalization, rectified linear units and convolution, respectively.
	
	On CIFAR-10, CIFAR-100 and SVHN datasets, the MixNets used in our experiments have three mixed link blocks with the same amount of layers. Before entering the first mixed link block, a convolution with $max(k_1, 2 \times k_2)$ output channels is performed on the input images. For convolutional layers with kernel size $3 \times 3$, each side of the inputs is zero-padded by one pixel to keep the feature-map size fixed. We use $1 \times 1$ convolution followed by $2 \times 2$ average pooling as transition layers between two contiguous  blocks. At the end of the last block, a global average pooling is performed and then a softmax classifier is attached. The feature-map sizes in the three  blocks are $ 32 \times 32$, $16 \times 16$, and $8 \times 8$, respectively. 
	We survey the network structure with three configurations: $\{L = 100, k_1 = 12, k_2 = 12\}$, $\{L = 250, k_1 = 24, k_2 = 24\}$ and $\{L = 190, k_1 = 40, k_2 = 40\}$ in practice.
	
	In our experiments on ImageNet dataset, we follow Arch-4 and use the network structure with four mixed link blocks on $224 \times 224$ input images. The initial convolution layer comprises $max(k_1, 2 \times k_2)$ filters whose size is $7 \times 7$ and stride is 2. The sizes of feature-maps in the following layers are determined by the settings of inner link parameter $k_1$ and outer link parameter $k_2$ (Table \ref{tab: nn-arch}), consequently.
	
	\section{Experiment}
	\label{sec:exp}
	In this section, we empirically demonstrate MixNet’s effectiveness and efficiency in parameter over the state-of-the-art architectures on many competitive benchmarks. 

\subsection{Datasets}
\noindent\textbf{CIFAR.} The two CIFAR datasets \cite{krizhevsky2009learning} consist of colored natural images with $32 \times 32$ pixels. CIFAR-10 consists of images drawn from 10 and CIFAR-100 from 100 classes. The training and test sets contain 50K and 10K images, respectively. We follow the standard data augmentation scheme that is widely used for these two datasets \cite{he2016deep,huang2016deep,lee2015deeply,romero2014fitnets,srivastava2015training,springenberg2014striving}. For preprocessing, we normalize the data using the channel means and standard deviations. For the final run we use all 5K training images and report the final test error at the end of training.

\noindent\textbf{SVHN.}
The Street View House Numbers (SVHN) dataset \cite{netzer2011reading} contains $32 \times 32$ colored digit images. There are 73,257 images in the training set, 26,032 images in the test set, and 531,131 images for extra training data. Following common practice \cite{Goodfellow2013Maxout,huang2016deep,lin2013network,sermanet2012convolutional}, We use all the training data (training set and extra training data)  without any data augmentation, and a validation set with 6,000 images is split from the training set. In addition, the pixel values in the dataset are divided by 255 and thus they are in the [0, 1] range. We select the model with the lowest validation error during training and report the test error.

\noindent\textbf{ImageNet.} The ILSVRC 2012 classification dataset \cite{deng2009imagenet} contains 1.2 million images for training, and 50K for validation, from 1K classes. We adopt the same data augmentation scheme for training images as in \cite{he2016deep,he2016identity}, and apply a single-crop with size $224 \times 224$ at test time. Following \cite{he2016deep,he2016identity}, we report classification errors on the validation set.

\subsection{Training}
All the networks are trained by using stochastic gradient descent (SGD). On CIFAR and SVHN we train using batch size 64 for 300 epochs. The initial learning rate is set to 0.1, and is divided by 10 at 50\% and 75\% of the total number of training epochs. On ImageNet, we train models with a mini-batch size 150 (MixNet-121) and 100 (MixNet-141) due to GPU memory constraints. To compensate for the smaller batch size, the models are trained for 100 epochs, and the learning rate is lowered by 10 times at epoch 30, 60 and 90. Following \cite{he2016deep}, we use a weight decay of $10^{-4}$ and a Nesterov momentum \cite{sutskever2013importance} of 0.9 without dampening. We adopt the weight initialization introduced by \cite{he2015delving}. {For the
the dataset without data augmentation (i.e., SVHN), we follow the DenseNet setting \cite{huang2016densely} and add a dropout layer \cite{srivastava2014dropout} after each convolutional layer (except the first one).}
\begin{table*}
	\scriptsize
	\centering
	\renewcommand\arraystretch{1.4}
	\newcommand{\tabincell}[2]{\begin{tabular}{@{}#1@{}}#2\end{tabular}}
	\caption{Error rates (\%) on CIFAR and SVHN datasets. $k_1$ and $k_2$ denote the parameters for inner and outer link modules, respectively. The best, second-best, and third-best accuracies are highlighted in red, blue, and green.}
	\vspace{-10pt}
	\begin{tabular}{l|cc|c|c|c}
		\hline
		Method & Depth & \#params & CIFAR-10 & CIFAR-100 & SVHN \\
		\hline
		RCNN-160 \cite{liang2015recurrent} & - & 1.87M & 7.09 & 31.75 & 1.80\\
		\hline
		DFN \cite{wang2016deeply} & 50 & 3.9M & 6.24 & 27.52 & - \\
		DFN-MR \cite{zhao2016connection}
		&50&24.8M&3.57&19.00&\textcolor{blue}{\bf1.55} \\
		\hline	
		FractalNet \cite{larsson2016fractalnet} &21 &38.6M&4.60&23.73&1.87\\
		\hline
		ResNet with Stochastic Depth \cite{huang2016deep} & 110 & 1.7M & 5.25 & 24.98 & 1.75 \\
		\hline
		ResNet-164 (pre-activation) \cite{he2016identity} & 164 & 1.7M & 4.80 & 22.11 & - \\
		ResNet-1001 (pre-activation) \cite{he2016identity} & 1001 & 10.2M & 4.92 & 22.71 & - \\
		\hline
		WRN-28-10 \cite{zagoruyko2016wide} & 28 & 36.5M & 4.00 & 19.25 & - \\
		\hline
		ResNeXt-29 ($8 \times 64$d) \cite{xie2017aggregated} & 29 & 34.4M & 3.65 & 17.77 & - \\ 
		ResNeXt-29 ($16 \times 64$d) \cite{xie2017aggregated} & 29 & 68.1M & 3.58 & 17.31 & - \\
		\hline
		DenseNet-100 ($k=24$) \cite{huang2016densely} & 100 & 27.2M & 3.74 & 19.25 & 1.59 \\
		DenseNet-BC-190 ($k=40$) \cite{huang2016densely} & 190 & 25.6M & 3.46 & \textcolor{green}{\bf17.18} & - \\
		\hline
		DPN-28-10 \cite{chen2017dual} & 28 & 47.8M & 3.65 & 20.23 & - \\
		\hline
		IGC-$L32M26$ \cite{zhang2017interleaved} &20&24.1M&\textcolor{blue}{\textbf{3.31}}&18.75&\textcolor{green}{\bf1.56}\\
		\hline
		MixNet-100 ($k_1=12, k_2=12$) & 100 & 1.5M & 4.19 & 21.12 & 1.57\\
		MixNet-250 ($k_1=24, k_2=24$) & 250 & 29.0M & \textcolor{green}{\bf 3.32} & \textcolor{blue}{\bf 17.06} & \textcolor{red}{\bf 1.51}\\
		MixNet-190 ($k_1=40, k_2=40$) & 190 & 48.5M & \textcolor{red}{\textbf{3.13}} & \textcolor{red}{\textbf{16.96}} & -\\
		\hline
	\end{tabular}
	\footnotetext[1]{\tiny Results from  https://github.com/Queequeg92/DualPathNet}
	\label{tab: res-cifar-fmnist}
	\vspace{-10pt}
\end{table*}

\begin{figure}
	\centering
	\setlength{\fboxrule}{0pt}
	\fbox{\includegraphics[width=0.44\textwidth]{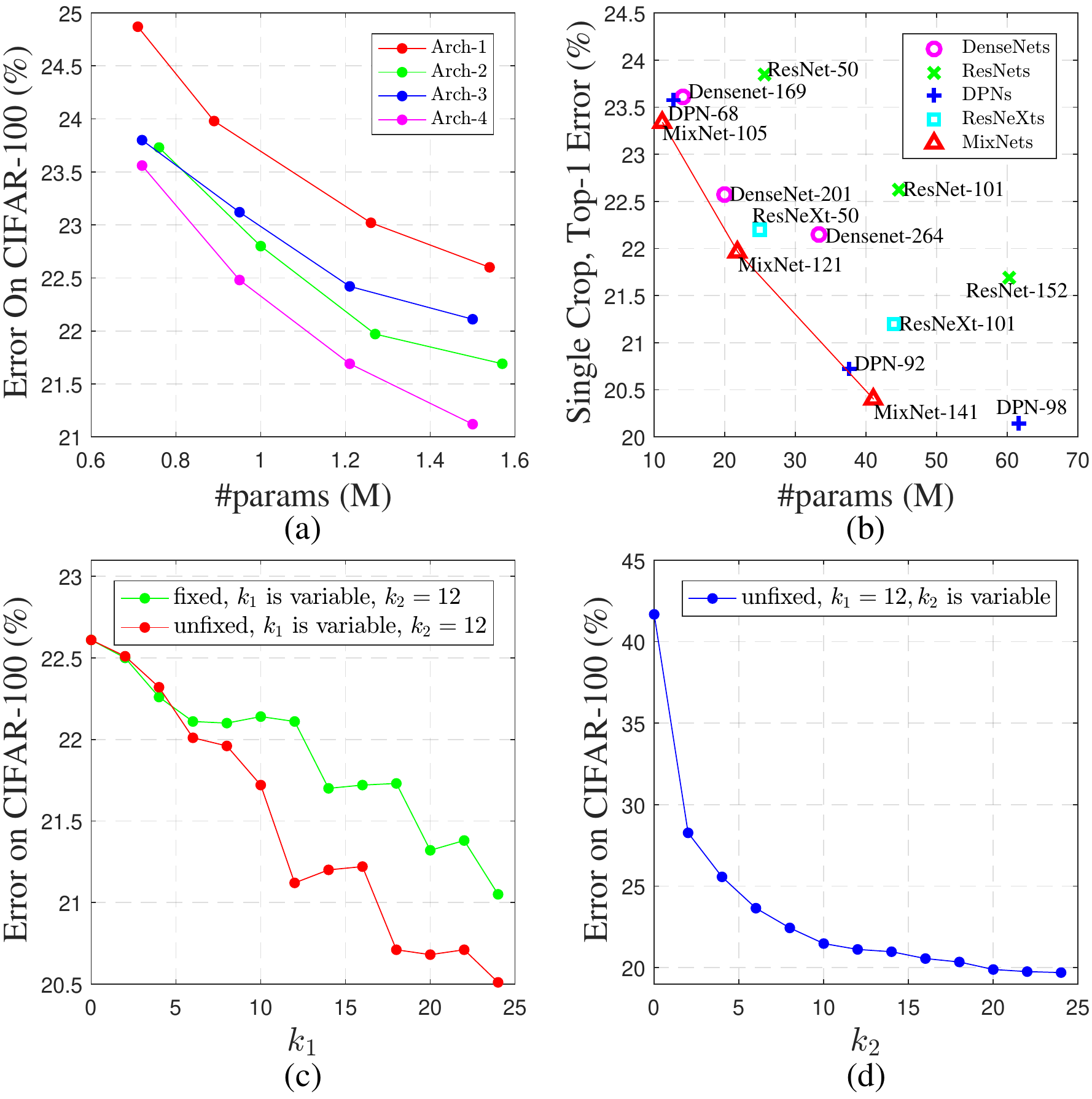}}
	\vspace{-10pt}
	\caption{The illustrations of the experimental results. (a) shows the parameter efficiency comparisons among the four architectures. {(b) is the comparison of the MixNets and the state-of-the-art architectures top-1 error (single-crop) on the ImageNet validation set as a function of model parameters.} (c) shows error rates of the models, whose inner link modules are fixed or unfixed. (d) shows error rates of the models with different outer link parameter $k_2$.}
	\label{fig:exp-stack-in-stack-out}
	\vspace{-10pt}
\end{figure}
\subsection{Ablation Study for Mixed Link Architecture}
\noindent\textbf{Efficiency comparisons among the four architectures.} We first evaluate the efficiency of the four representative architectures which are derived from the mixed link architecture. The comparisons are based on various amount of parameters (\#params). Specifically, we increase the complexities of the four architectures in parallel and evaluate them on CIFAR-100 dataset. The experimental results are reported in Fig. \ref{fig:exp-stack-in-stack-out} (a), from which we can find that with various similar parameters, Arch-4 outperforms all other three architectures by a margin. It demonstrates the superior efficiency in parameter of Arch-4 which is exactly used in our proposed MixNets.

\noindent\textbf{Fixed \emph{vs.} unfixed} for the inner link modules. Next we investigate ``which is the more effective setting for the inner link modules -- fixed or unfixed?''. To ensure a fair comparison, we hold the outer link parameter $k_2$ constant and train MixNets with different inner link parameter $k_1$. In details, we set $k_2$ to $12$, and let $k_1$ increase from $0$ to $24$. The models are also evaluated on CIFAR-100 dataset. Fig. \ref{fig:exp-stack-in-stack-out} (c) shows the experimental results, from which we can find that with the growing of $k_1$, the test error rate keeps dropping. Furthermore, with the same inner link parameter $k_1$, the models with unfixed inner link modules (red curve) have much lower test errors than the models with the fixed ones (green curve), which suggests the superiority of unfixed inner link module.

\noindent\textbf{Outer link size.} We then study the effect of outer link size $k_2$ by setting $k_1 = 12$, under the configurations with the effective unfixed inner link modules on CIFAR-100 dataset. Fig. \ref{fig:exp-stack-in-stack-out} (d) illustrates that the increasement of $k_2$ reduces the test error rate consistently. However, the performance gain becomes tiny when $k_2$ is relatively large. 



\subsection{Experiments on CIFAR and SVHN}
We train MixNets with different depths $L$, inner link parameters $k_1$ and outer link parameters $k_2$. The main results on CIFAR and SVHN are shown in Table \ref{tab: res-cifar-fmnist}. Except for DPN-28-10 which is from  https://github.com/Queequeg92/DualPathNet, all other reported results are directly borrowed from their original papers.

As can be seen from the bottom rows of Table \ref{tab: res-cifar-fmnist}, MixNet-190 outperforms many state-of-the-art architectures consistently on CIFAR datasets. Its error rates, $3.13\%$ on CIFAR-10 and $16.96\%$ on CIFAR-100, are significantly lower than the error rates achieved by DPN-29-10. Our results on
SVHN are even more encouraging. MixNet-100 achieves comparable test errors with DFN-MR (24.1M) and IGC-$L32M26$ (24.8M) whilst costing only 1.5M parameters. 

\subsection{Experiments on ImageNet}

\begin{table}
	\scriptsize
	\centering
	\renewcommand\arraystretch{1.4}
	\newcommand{\tabincell}[2]{\begin{tabular}{@{}#1@{}}#2\end{tabular}}
	\caption{{The top-1 and top-5 error rates on the ImageNet validation set, with single-crop / 10-crop testing.}}
	\vspace{-8pt}
	\begin{tabular}{l|c|c|c}
	\hline
	Method & \#params & top-1 & top-5 \\
	\hline
	ResNet-101 \cite{he2016deep} & 44.55M & 22.6 & 6.4 \\
	ResNet-152 \cite{he2016deep} & 60.19M & 21.7 & 6.0 \\
	\hline
	DenseNet-169 \cite{huang2016densely} & 14.15M & 23.8 & 6.9 \\
	DenseNet-264 \cite{huang2016densely} & 33.34M & 22.2 & 6.1 \\
	\hline
	ResNeXt-50 (32 $\times$ 4d) \cite{xie2017aggregated} & 25M & 22.2 & - \\
	ResNeXt-101 (32 $\times$ 4d) \cite{xie2017aggregated} & 44M & 21.2 & 5.6 \\
	\hline
	DPN-68 (32 $\times$ 4d) \cite{chen2017dual} & 12.61M & 23.7 & 7.0 \\
	DPN-92 (32 $\times$ 3d) \cite{chen2017dual} & 37.67M & 20.7 & 5.4 \\
	DPN-98 (32 $\times$ 4d) \cite{chen2017dual} & 61.57M & 20.2 & 5.2 \\
	\hline
	MixNet-105 ($k_1=32, k_2=32$) & 11.16M & 23.3 & 6.7 \\
	MixNet-121 ($k_1=40, k_2=40$) & 21.86M & 21.9 & 5.9 \\
	MixNet-141 ($k_1=48, k_2=48$) & 41.07M & 20.4 & 5.3 \\
	\hline
	\end{tabular}
	\label{tab: res-imagenet}
	\vspace{-10pt}
\end{table}

We evaluate MixNets with different depths and inner/outer link parameters on the ImageNet classification task, and compare it with the representative state-of-the-art architectures. 
We report the single-crop and 10-crop validation errors of MixNets on ImageNet in Table \ref{tab: res-imagenet}. The single-crop top-1 validation errors of MixNets and different state-of-the-art architectures as a function of the number of parameters are shown in Fig. \ref{fig:exp-stack-in-stack-out} (b). The results reveal that MixNets perform on par with the state-of-the-art architectures, whilst requiring significantly fewer parameters to achieve better or at least comparable performance. For example, MixNet-105 outperforms DenseNet-169 and DPN-68 with only 11.16M parameters. MixNet-121 (21.86M) yields better validation error than ResNeXt-50 (25M) and Densenet-264 (33.34M). Furthermore, the results of MixNet-141 are very close to the ones of DPN-98 with 50\% fewer parameters.

\section{Conclusion}
\label{sec:con}
In this paper, we first prove that ResNet and DenseNet are essentially derived from the same \emph{fundamental} ``dense topology'', whilst their only difference lies in the specific form of connection. Next, basing on the analysis of superiority and insufficiency of their distinct connections, we propose a highly efficient form of it -- the Mixed Link Networks (MixNets), whose connection combines both flexible inner link modules and outer link modules. Further, MixNet is a generalized structure that ResNet, DenseNet and DPN can be regarded as special cases of it. Extensive experimental results demonstrate that our proposed MixNet is efficient in parameter.



\bibliographystyle{named}
\bibliography{ijcai18}

\end{document}